\pdfoutput=1

\documentclass[11pt]{article}

\usepackage{ACL2023}
\usepackage[T1]{fontenc}
\usepackage[utf8]{inputenc}
\usepackage{microtype}
\usepackage{inconsolata}
\usepackage{float}              
\usepackage{subfig}             
\usepackage{overpic}            
\usepackage{multirow}
\usepackage{times} 
\usepackage{helvet} 
\usepackage{courier} 
\usepackage{graphicx}
\usepackage{bm}
\usepackage{natbib}  
\usepackage{caption} 
\usepackage{algorithm}
\usepackage{algorithmic}
\usepackage{amsthm}
\usepackage{amsmath}
\usepackage{multirow}
\usepackage{booktabs}
\usepackage{newfloat}
\usepackage{listings}
\usepackage{bibentry}
\usepackage[skins]{tcolorbox} 
\newtheorem{theorem}{Theorem}

\newtheorem{corollary}{Corollary}
\newtheorem{definition}[theorem]{Definition}
\newtheorem{assumption}[theorem]{Assumption}

\usepackage{xcolor}

\title{The Scaling Law for LoRA Base on Mutual Information Upper Bound}

\author{
    Jing Zhang,
     Hui Gao,
    Peng Zhang\thanks{Corresponding author: Peng Zhang (pzhang@tju.edu.cn)},
    Shuzhen Sun,
    Chang Yang,
    Yuexian Hou\thanks{Corresponding author: Yuexian Hou (yxhou@tju.edu.cn)}\\
    College of Intelligence and Computing, Tianjin University, Tianjin, China\\
    \{zhang\_jing@tju.edu.cn, pzhang, yxhou\}@tju.edu.cn\\
}

\begin{document}
\maketitle

\begin{abstract}
LoRA (Low-Rank Adaptation) is a widely used model fine-tuning method. In fine-tuning, the law among model performance, model parameters, and data complexity has been a focal issue in the field. Existing methods often leverage external metrics (such as cross-entropy or perplexity) to evaluate model performance. In the fine-tuning process for large models, two types of knowledge are typically involved: the frozen, general knowledge acquired by the model during pre-training and the new knowledge learned through the LoRA module from the current data. Generally, the less LoRA's learned knowledge relies on the large model, the more it captures the specific knowledge of new data, thereby enhancing its adaptability to new tasks. However, external metrics do not readily capture the dependency relationship between these two types of knowledge. Therefore, we designed an internal metric based on the \textbf{\textit{Mutual Information Upper Bound (MIUB)}} theory to investigate the scaling law of large-model LoRA fine-tuning. In our experiments, we validated this approach on benchmark datasets, using the Llama3-8B and Phi3-3B models. The results show that the proposed MIUB metric aligns more accurately and stably with the scaling law of LoRA fine-tuning compared to cross-entropy and perplexity.
\end{abstract}

\section{Introduction}

Pre-trained on vast amounts of data, large language models like GPT-X~\cite{achiam2023gpt} and LLaMA3~\cite{dubey2024llama} have achieved remarkable results in general domains. However, to address various personalized needs, especially under the pressure of inference deployment costs, fine-tuning serves as an effective method for model compression, enhancing the model's personalization and multi-tasking capabilities with relatively small datasets~\cite{kim2024memory,wang2023cocktailsgd, ge2023context}. Among these, LoRA (Low-Rank Adaptation)~\cite{hu2021lora, yang2024moral} fine-tuning leverages the idea of low-rank approximation. By freezing the parameters of the large model, it only uses a small number of newly added low-rank parameter matrices to learn the structured knowledge in the new data.

In order to reduce the blindness of large model pre-training, some work has proposed that there is a scaling law for large model pre-training~\cite{kaplan2020scaling,wei2024large}, that is, as the size of the large model increases and the amount of pre-training data increases, the effect of pre-training usually changes regularly. However, there is currently no systematic research on the scaling law of large model LoRA fine-tuning, which will undoubtedly increase the computational cost and trial-and-error cost of fine-tuning.

The factors that affect the effect of model fine-tuning mainly include large model size, the size of LoRA parameters, and data size. However, how to accurately quantify the impact of these factors on the effect is still an urgent problem to be solved. From a general perspective, the effect after fine-tuning is mainly related to two parts of knowledge, one is the meta-knowledge relied on from the large language model, and the other is the generalized knowledge learned by the newly added parameters~\cite{mao2024survey, jovanovic2024trends}.  Some work has shown that when fine-tuning large models, there will be conflicts between new and old knowledge~\cite{shi2024ircan}.The existing evaluation metrics, such as test set PPL (perplexity) and cross-entropy, primarily focus on assessing the overall distribution of the model. However, they have yet to fine-grainedly quantify the relationship between the feature distribution of the base LLM and the feature distribution introduced by the LoRA module under the fine-tuning paradigm. Therefore, some work also shows that evaluations based on indicators such as Cross-Entropy and PPL are sometimes not accurate and stable~\cite{wei2024large}.

In order to measure the relationship between the feature distribution space of the large model and the feature distribution space of LoRA, we propose to use a mutual information metric with an upper bound to accurately measure the impact of model size and data scale on the LLM effect. This paper adds the LoRA structure to the attention layer and the FFN layer. Relying on the structural advantages of LoRA, it concisely and effectively calculates the Mutual Information Upper Bound (MIUB)  between the output distribution of the large model and the output distribution of LoRA after residual connection.

In general, the Mutual Information Upper Bound (MIUB) between the hidden representation of the large model and that of the LoRA module can serve as a quality indicator for evaluating the fine-tuning effect, particularly in the context of model compression and personalization. Experimental results show that the MIUB decreases with the increase of the size of the large model, decreases with the increase of LoRA rank, and also decreases with the increase of data size (i.e., length or complexity). Furthermore, the upper bound of mutual information not only conforms to the scaling laws, but is also more stable and more in line with the changing trend of the actual effect of the model (such as Accuracy) than traditional outward evaluation indicators. In addition, compared with the prompt template used for fine-tuning, this article also shows the size rules of MIUB under different prompt templates.

\begin{itemize}
    \item Building on the LoRA fine-tuning framework, we propose an evaluation metric based on the upper bound of mutual information between the hidden distributions of the large model and the LoRA model, aimed at quantifying the performance of large models under model compression and personalized fine-tuning requirements.
    \item Building on the upper bound of mutual information, this paper systematically studies the scaling laws of LoRA fine-tuning based on the analysis of internal model relationships for the first time.
    \item Experimental results show that the upper bound of mutual information not only conforms to scaling laws in terms of model size and data complexity, but its results are more accurate and stable than traditional Cross-Entropy and PPL indicators.
    
    \end{itemize}

\section{Related Works}

\subsection{Low-Rank Adaptation}
Low-Rank Adaptation (LoRA) is a method for effectively fine-tuning pre-trained models by introducing low-rank matrices while keeping the large model parameters frozen. In recent years, it has become a primary technique for fine-tuning large models, offering high flexibility and efficiency. ~\citep{hu2021lora} was the first to propose the application of LoRA in large models, with experiments showing that it can achieve good fine-tuning results with a small number of additional parameters, reducing both training parameters and computational overhead. The core idea of this method is to decompose the weight updates of the model into low-rank matrices, significantly reducing computational costs while maintaining model performance. In recent years, various works have focused on reducing the cost of model fine-tuning and enhancing its generalization capabilities in the design of LoRA structures. \citep{ding2023sparse} further reduced the computational cost of LoRA by using gating units to dynamically adjust the intrinsic rank. Additionally, combining LoRA with MoE techniques has also provided assurance for enhancing its generalization ability. LoRAHub~\cite{huang2023lorahub} selects different LoRA combinations for task generalization. \citep{dou2024loramoe}  proposed MoELoRA, which utilizes both LoRA and MoE for specific task adjustment and multi-task processing. ~\citep{liu2023moelora} introduced the multimodal learning capabilities of multimodal expert models.

\subsection{Scaling Law}

Scaling laws have been a persistent topic in both nature and science~\cite{gan2021universal}, and in recent years, they have also shown strong guiding capabilities in the field of Large Language Models (LLMs). In the field of neural networks, scaling laws critically demonstrate how model performance scales with increases in computational resources, data, and model parameters. \citep{hu2021lora} first introduced the ``scaling law'' for neural language models, indicating that larger models trained on more data tend to perform better. \citep{zhang2024scaling} comprehensively tested the scaling laws of fine-tuning frameworks under existing evaluation metrics. In the information retrieval domain, \citep{fang2024scalinglawsdenseretrieval} proposed using contrastive log-likelihood as a metric to assess whether retrieval models adhere to scaling laws. In the compression domain, ~\citep{wei2024large} introduced the information-theoretic Matrix entropy to measure the performance of large models, showing that Matrix entropy is more accurate and stable compared to the unstable CE (cross-entropy) and PPL (perplexity) metrics. This work has inspired us to investigate the internal relationships within models to systematically measure the performance of LoRA in compression and multi-task scenarios.

\begin{figure*}[ht]
    \centering
    \includegraphics[width=0.9\textwidth]{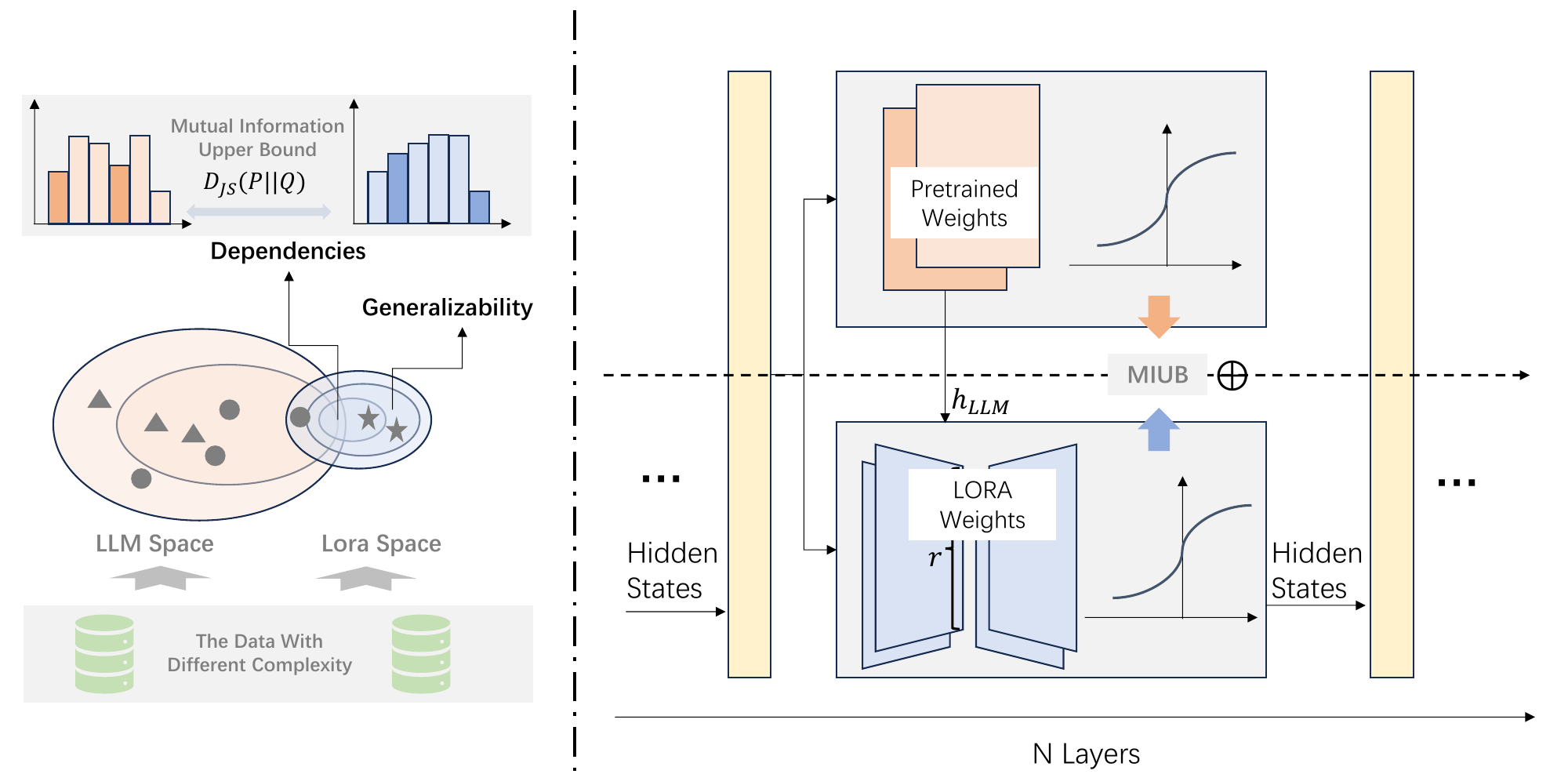} 
    \caption{The overall method}
    \label{fig1}
\end{figure*}

\section{Methodology}

\subsection{The Scaling Law of LoRA}

For newly added fine-tuning data, without disrupting the feature space of the large model itself (i.e., by freezing the parameters of the large model), LoRA relies on some of the meta-knowledge of the LLM and learns new specific features by adding low-rank parameter weights. Therefore, as shown in Figure~\ref{fig1}, there is a natural dependence and generalization relationship between the LLM and LoRA modules. Furthermore, we model the dependency relationship between them as mutual information.

\begin{assumption}
In neural networks, the dependency between large language models (LLM) and LoRA is manifested through the intertwining of their feature spaces, which can be quantified by mutual information. This mutual information not only measures the information obtained about the distribution of LoRA from the LLM variables but also reveals the extent of their overlap in feature space. Under this dependency, the feature distribution of LoRA can be viewed as conditional, co-evolving with the distribution of the LLM in the hidden space.
\end{assumption}

\begin{definition}
Formally, let $\bm{O}$ represent the hidden states from the feature space of the LLM, and $\bm{L}$ represent the hidden states from the feature space of LoRA. The mutual information $\mathcal{I}(\bm{O}; \bm{L})$ between $\bm{O}$ and $\bm{L}$ is defined as:

\begin{equation}
\mathcal{I}(\bm{O}; \bm{L}) = \iint_{\bm{O} \times \bm{L}} p(\bm{o}, \bm{l}) \log \left( \frac{p(\bm{o}, \bm{l})}{p(\bm{o}) p(\bm{l})} \right) \, \mathrm{d}\bm{o} \, \mathrm{d}\bm{l},
\end{equation}

where $p(\bm{o}, \bm{l})$ is the joint probability density function of $\bm{O}$ and $\bm{L}$, and $p(\bm{o})$ and $p(\bm{l})$ are the marginal probability density functions of $\bm{O}$ and $\bm{L}$, respectively. This double integral defines mutual information over the continuous domain, with every point in the feature space contributing to this information measure. Furthermore, a higher value of $\mathcal{I}(\bm{O}; \bm{L})$ indicates a stronger dependency between $\bm{O}$ and $\bm{L}$ and a greater degree of shared information.      
\end{definition}

The mutual information measure based on the LoRA architecture quantifies the role of the large model's feature space information in reducing the uncertainty of the LoRA distribution's feature space, thereby capturing the dependency between the two. However, a strong dependency does not necessarily align with the actual performance of the model, and it is difficult to find a consistent pattern across models of different scales. Therefore, this paper introduces an upper bound on mutual information to measure the distance between the large model distribution and the LoRA distribution.

\begin{theorem}[Upper Bound of Mutual Information Based on JS Divergence]
Let \( \bm{O} \) and \( \bm{L} \) be random variables in the original LLM and LoRA probability spaces with joint distribution \( \bm{P}_{OL} \) and marginal distributions \( \bm{P}_O \) and \( \bm{P}_L \), respectively. Let \( \mathcal{D}_{\text{JS}}(\bm{P}_O \| \bm{P}_L) \) denote the Jensen-Shannon divergence between \( \bm{P}_O \) and \( \bm{P}_L \). The mutual information \( \mathcal{I}(\bm{O}; \bm{L}) \) satisfies the following upper bound:

\begin{equation}
\mathcal{I}(\bm{O}; \bm{L}) \leq \log(2) \cdot \mathcal{D}_{\text{JS}}(\bm{P}_{OL} \| \bm{P}_O \bm{P}_L)
\end{equation}
where the Jensen-Shannon divergence between two distributions \( \bm{P} \) and \( \bm{Q} \) is defined as:
\begin{equation}
\mathcal{D}_{\text{JS}}(\bm{P} \| \bm{Q}) = \frac{1}{2} D_{\text{KL}}(\bm{P} \| \bm{M}) + \frac{1}{2} D_{\text{KL}}(\bm{Q} \| \bm{M}),
\end{equation}
and \( D_{\text{KL}} \) denotes the Kullback-Leibler divergence, with \( \bm{M} = \frac{1}{2} (\bm{P} + \bm{Q}) \), where \( \bm{P} \) and \( \bm{Q} \) are the distributions being compared.

\begin{proof} 
We begin by noting that mutual information \( \mathcal{I}(\bm{O}; \bm{L}) \) can be expressed as the Kullback-Leibler divergence between the joint distribution \( \bm{P}_{OL} \) and the product of the marginal distributions \( \bm{P}_O \) and \( \bm{P}_L \):
\begin{equation}
\mathcal{I}(\bm{O}; \bm{L}) = D_{\text{KL}}(\bm{P}_{OL} \| \bm{P}_O \bm{P}_L).
\end{equation}

The Jensen-Shannon divergence between the joint distribution \( \bm{P}_{OL} \) and the product of the marginal distributions \( \bm{P}_O \bm{P}_L \) is defined as:
\begin{equation}
\mathcal{D}_{\text{JS}}(\bm{P}_{OL} \| \bm{P}_O \bm{P}_L) = \frac{1}{2} D_{\text{KL}}(\bm{P}_{OL} \| \bm{M}) + \frac{1}{2} D_{\text{KL}}(\bm{P}_O \bm{P}_L \| \bm{M}),
\end{equation}
where \( \bm{M} = \frac{1}{2} (\bm{P}_{OL} + \bm{P}_O \bm{P}_L) \). This form is symmetric, and the divergence is bounded below by the mutual information \( \mathcal{I}(\bm{O}; \bm{L}) \). Specifically, we have the following inequality:
\begin{equation}
\mathcal{D}_{\text{JS}}(\bm{P}_{OL} \| \bm{P}_O \bm{P}_L) \geq \frac{1}{2} \mathcal{I}(\bm{O}; \bm{L}).
\end{equation}

By multiplying both sides by \( \log(2) \), we obtain the desired upper bound:
\begin{equation}
\mathcal{I}(\bm{O}; \bm{L}) \leq \log(2) \cdot \mathcal{D}_{\text{JS}}(\bm{P}_{OL} \| \bm{P}_O \bm{P}_L).
\end{equation}
Therefore, the upper bound of mutual information is connected to the Jensen-Shannon divergence in the context of the LoRA architecture.
\end{proof}

\end{theorem}
By introducing the upper bound of mutual information to measure dependence within the LoRA architecture, it ensures that the dependence will not decrease indefinitely during the data measurement process, but will instead stabilize within a certain range and approach its upper bound. This approach provides a theoretical upper limit for the dependence in the LoRA architecture, while also guaranteeing its convergence, ensuring that the dependence ultimately stabilizes within a finite range and avoiding excessive fluctuations or infinite reduction.

\begin{corollary}
Here is a scaling law that focuses on the model size, LoRA rank size, and dataset size during LoRA fine-tuning:

\begin{equation}
\text{MIUB}(N, R, D) = A \left( \frac{N_0}{N} \right)^\alpha + B \left( \frac{R_0}{R} \right)^\beta + C \left( \frac{D_0}{D} \right)^\gamma
\end{equation}
where $\text{MIUB}(N, R, D)$ is the metric as a function of the number of parameters in the large model $N$, the LoRA rank size $R$, and the dataset size $D$. $N_0, R_0, D_0$ are scaling constants that normalize the respective terms. $\alpha, \beta, \gamma$ are scaling exponents that describe how the MIUB scales with respect to the model size, LoRA rank size, and dataset size, respectively. $A, B, C$ are constants that depend on the specific problem and architecture.
\end{corollary}

\begin{corollary}

As shown in Figure~\ref{fig1}, supported by a vast amount of data, the large model is able to learn rich meta-knowledge. During the fine-tuning process with new data, the LoRA module inevitably relies on the meta-knowledge of the large model when learning knowledge from the new data. Therefore, compared to traditional metrics such as Loss and ACC, the upper bound of mutual information calculated from the hidden distributions of the large model and LoRA, as presented in this paper, provides a more comprehensive measure of the dependency between them. When the upper bound of mutual information is larger, it indicates a stronger dependency between the two, suggesting that the LoRA module has learned less domain-specific knowledge from the new data and performs worse on the fine-tuning data.

\end{corollary}

\subsection{The FrameWork of Fine-tuning and Evaluation}

The paper adds LoRA structures to all the Dense Linear layers in the Attention and FFN modules of a large model. The original parameters of the large model are frozen, and only the LoRA components are trained during fine-tuning. Specifically, the hidden states of the large model are denoted as $h_{LLM}^{m}$, and the hidden states of LoRA, $h_{lora}^{m}$, are obtained by adding the hidden states of the large model to the output of LoRA.

The hidden states $h_{LLM}^{m}$ and $h_{LoRA}^{m}$ are converted into probability distributions using the softmax function. Then, the upper bound of the mutual information (MIUB) between these two probability distributions is calculated, as shown in Figure~\ref{fig1}. By summing the MIUBs of all the LoRA components, we obtain the MIUB for a single sample. The average MIUB across all samples gives the final evaluation value:
\begin{equation}
M = \frac{1}{N} \sum_{\omega_{\text{set}}} \sum_{m} D_{JS}^{m}(P \| Q)
\end{equation}

where $D_{JS}^{m}(P || Q)$ represents the Jensen-Shannon divergence between the probability distributions $P$ and $Q$ for the $m$-th component.

\begin{tcolorbox}
[colback=gray!10, colframe=black, 
colbacktitle=gray, coltitle=white,title=\textbf{Prompt}]
\textbf{Train Prompt}\\
Choose the correct answer for the following question: $x_i$\\
Answer: $y_i$\\
\textbf{Test Prompt 1}\\
Choose the correct answer for the following question. Here is an example shown below: $s_i$\\
The new question is: $x_i$\\
Answer: $y_i$

\textbf{Test Prompt 2}\\
Choose the correct answer for the following question. Here is an positive example shown below: $s_i$\\
Here is an negative example shown below:  $n_i$\\
The new question is: $x_i$\\
Answer: 

\textbf{Test Prompt 3}\\
Please note that you can only choose from A, B, c or D. Choose the correct answer for the following question: $x_i$\\
Answer: 

\label{pro:1}
\end{tcolorbox}
As shown in the figure, we employed prompt learning during the fine-tuning of the large model. Taking a classification task as an example, the Train Prompt instructs the model to select the correct option and serves as a zero-shot template. During testing, in addition to the zero-shot prompt, we also have the option to use a 1-shot template (Test Prompt 1), which includes one positive and one negative example, a few-shot template, and a template that imposes restrictions on the task output (Test Prompt 3). We also evaluate the model's performance across different prompt templates.

\begin{table*}[h]
    \centering
    \caption{The Main Results. Evaluate the effect of model testing with changing  LoRA rank and LLM size based on compression}
    \resizebox{0.80\textwidth}{!}{
    \begin{tabular}{c|c|c|ccc|cccc}
        \toprule
         Dataset&Model & Matrices &  32 & 128& 512 &$share_8$ &$share_4$ &$share_2$ &$share_1$\\
         \midrule

         \multirow{6}{*}{ARCE}  &\multirow{3}{*}{Phi3}  & ACC  & 0.951  &0.951  &0.969 & 0.955 & 0.949  &0.951 & 0.955 \\
               &  & CE  & 0.018   & 0.162  & 0.007  &  23.977& 9.647 &2.514  & 0.077 \\
               &  & MIUB  &1586.0    & 1566.6 & 1567.7 & 1902.6 &1712.6 & 1643.0 &  1597.3 \\
                    \cmidrule{2-10}

                &\multirow{3}{*}{LLaMA3}  
                & ACC  &0.912 &0.809 &0.862 &0.135  &0.901  & 0.894    &0.912  \\
                &    & CE  & 14.339 &13.860  &13.427 &16.013    & 11.594  &13.399 & 14.339\\
                 &   & MIUB & 3405.1 &3397.5 & 3398.8 &3779.9.2  &3579.5    &  3443.7 &3405.1\\

\midrule
         \multirow{6}{*}{ARCC}  &\multirow{3}{*}{Phi3}  & ACC &0.859  & 0.863  &0.887  & 0.849 & 0.863 & 0.887 & 0.873 \\
               &  & CE & 0.018  &0.159  &0.007  &24.661  & 9.606 & 2.510 &  0.070\\
               &  & MIUB  &1596.2   &  1579.1  & 1578.3 & 1913.3 & 1728.5 & 1657.4 &1607.6 \\
                    \cmidrule{2-10}
                &\multirow{3}{*}{LLaMA3}  
                & ACC   &0.823 &  0.720 & 0.792 &0.078 & 0.154 &0.816 &0.823  \\
                &    & CE  &13.463 &12.798 &12.342 &15.752  & 13.328 &11.295 & 13.463 \\
                 &   & MIUB &3446.6 &3435.1  &3439.3    &3808.4  & 3613.7  &   3485.0  & 3446.6\\

\midrule
         \multirow{6}{*}{Hel}  &\multirow{3}{*}{Phi3}  & ACC & 0.809  &0.796  &0.814  & 0.789 &  0.809 & 0.789 &0.813 \\
               &  & CE  & 0.009  & 0.191 & 0.014 & 18.215 & 14.896 & 0.057 & 0.007 \\
               &  & MIUB &1521.4   & 1518.7 & 1511.6 & 1884.9 & 1685.9 & 1608.0 & 1542.1 \\
                    \cmidrule{2-10}
                &\multirow{3}{*}{LLaMA3}  
                & ACC    &  0.882 &0.882& 0.875 & 0.856 & 0.817 &0.872 &  0.882 \\
                &    & CE &0.014 & 0.020 & 0.010  &12.710  & 7.447 & 2.124 &0.014\\
                 &   & MIUB &3282.0  &3265.1 &3258.3 &3706.6  & 3475.7  & 3282.1& 3282.0\\

\midrule

         \multirow{6}{*}{PIQA}  &\multirow{3}{*}{Phi3}  & ACC  &0.848 &0.849 & 0.836 & 0.852  & 0.859 & 0.858 &0.863 \\
               &  & CE &0.020   & 0.362 & 0.012 & 17.117 & 13.471 & 0.494  &0.065\\
               &  & MIUB  &1610.5   &  1590.2 &  1586.4  &1906.6  &1742.5  & 1693.0 &1621.0 \\
                    \cmidrule{2-10}
                &\multirow{3}{*}{LLaMA3}  
                & ACC  &0.891 & 0.897 &0.900 &0.885  &  0.837& 0.891  & 0.891 \\
                &    & CE  &10.299  &13.514  & 12.947  &28.504 & 12.926  &12.793 &10.299\\
                 &   & MIUB&3497.6 & 3489.1 &3485.9  &3837.0  &  3658.6 &3540.3 &  3497.6 \\
\midrule

         \multirow{6}{*}{Win}  &\multirow{3}{*}{Phi3}  & ACC  &  0.815  & 0.822 & 0.814 & 0.817 & 0.803 &0.821 &0.830 \\
               &  & CE  & 0.024  & 0.242 & 2.452 &23.470 &  11.055 & 9.077  & 0.004 \\
               &  & MIUB  & 1543.6 & 1530.5 & 1531.4 & 1883.2 & 1679.8 & 1616.7 &1557.5 \\
                    \cmidrule{2-10}
    
                &\multirow{3}{*}{LLaMA3}  
                & ACC  & 0.863 & 0.858 & 0.866 & 0.832 & 0.725  &  0.855  & 0.863 \\
                &    & CE &  4.700  &6.462  &4.984  &12.109  & 9.211  & 7.702 & 4.700\\
                 &   & MIUB &3233.0 & 3232.7  & 3217.7 &3698.0  &  3407.2 &  3238.6  & 3233.0\\

\midrule

     \multirow{6}{*}{AVG}  &\multirow{3}{*}{Phi3}  & ACC  &  0.856  &0.856\textcolor{black}{$\uparrow$}  & 0.864\textcolor{black}{$\uparrow$} & 0.852  &0.857{$\uparrow$} & 0.861{$\uparrow$} &0.867{$\uparrow$} \\
           &  & CE  & 0.018   & 0.251\textcolor{red}{$\uparrow$} & 0.602 \textcolor{red}{$\uparrow$} &20.110  & 12.695 \textcolor{green}{$\downarrow$}  & 2.763\textcolor{green}{$\downarrow$}  &0.035\textcolor{green}{$\downarrow$}  \\
           &  & MIUB  & \textbf{1564.6}  & \textbf{1551.5}\textcolor{green}{$\downarrow$} & \textbf{1548.6}\textcolor{green}{$\downarrow$} & \textbf{1894.5} & \textbf{1706.7}\textcolor{green}{$\downarrow$} & \textbf{1642.2}\textcolor{green}{$\downarrow$} & \textbf{1579.1}\textcolor{green}{$\downarrow$}\\
                \cmidrule{2-10}
            &\multirow{3}{*}{LLaMA3}  
            & ACC   & 0.874  &0.833 &0.859 &0.557  &0.550 & 0.867  &0.874  \\
            &    & CE   & 5.803&6.978 & 6.446 & 17.363  & 10.181  & 7.544   & 5.803\\
             &   & MIUB & \textbf{3350.2}   & \textbf{3340.2}&\textbf{3334.0} & \textbf{3752.3} & \textbf{3527.8}  &  \textbf{3368.6} &\textbf{3350.2}   \\

        \bottomrule
    \end{tabular}
    }
    \label{tab:my_label}
\end{table*}

\section{Experiments}
In this section, we will evaluate the proposed model structure on natural language tasks and verify the effectiveness of various measures we use. All experiments were performed on NVIDIA A-800 GPU.
\subsection{Model and Hyperparameters}

We used Llama3~\cite{dubey2024llama} and Phi3-3B~\cite{abdin2024phi} as our testing model, Llama3 has 8B parameters and 32 layers and Phi3 has 3B parameters and 32 layers, we use them to test the best model settings on models of different sizes. We use Adam as the optimizer with a learning rate of $4 \times 10^{-5}$ for fine-tuning downstream tasks and set the batch size to 32.
\subsection{Dataset}

We used our proposed  structure on five popular zero-shot generation tasks, including PIQA~\citep{bisk2020piqa}, ARC-Challenge~\citep{clark2018think}, ARC-Easy~\citep{clark2018think}, Winogrande~\citep{sakaguchi2021winogrande}, and HellaSwag~\citep{zellers2019hellaswag}, with higher accuracy, indicating that Mooe has a stronger parameter fine-tuning ability to handle downstream tasks.

For perplexity verification, we chose Wiki2~\cite{merity2016pointer} and PTB~\cite{marcus1994penn} as our verification data sets. Lower perplexity indicates that the compressed model has a stronger ability to maintain the output distribution of the original model.

\subsection{Metrics}

The test metrics used in this paper are as follows:

\textbf{Cross Entropy Loss (CE)} measures how well the predicted probability distribution \( q \) approximates the true distribution \( p \). Lower cross-entropy indicates better predictive performance.
\begin{equation}
H(p, q) = -\sum_{i=1}^N p(w_i) \log q(w_i)
\end{equation}

where $p(w_i)$ is true probability distribution of the $i$-th event,  $q(w_i)$ is predicted probability distribution of the $i$-th event,  and $N$ is number of events or classes.

\textbf{Perplexity (PPL)} is the exponentiated average negative log-likelihood of a sequence. It measures how well a language model predicts a sequence, with lower values indicating better performance.
\begin{equation}
\text{PPL} = \exp\left(-\frac{1}{N} \sum_{i=1}^N \log q(w_i)\right)
\end{equation}

where  $q(w_i)$ is the predicted probability of the $i$-th word and $N$ represents number of words in the sequence.

\textbf{Mutual information} measures the amount of information shared between LLM space and LoRA Space.

\subsection{Scaling Law Setting}

\begin{itemize}
    \item 
    For the scaling settings of the LoRA components, we primarily adjusted the rank to different sizes, specifically {32, 128, 512}.
    \item 
    For the large model, we applied a parameter-sharing compression method to adjust the scaling of the model. To ensure that the model's basic performance is not unfairly affected or that abnormal experimental results do not occur due to compression, we fixed the first 16 layers of the Phi3 and llama3 models and applied different parameter-sharing strategies to the last 16 layers: sharing every eight layers ($share_8$), every four layers ($share_4$), every two layers ($share_2$), and no sharing ($share_1$).
    \item 
    In the data scaling section, we selected 100 data samples from each of the test sets across multiple tasks, with data lengths in the ranges of [1, 100], [101, 200], and [201, 300].
\end{itemize}

\subsection{Main Results}
We conducted experiments on five benchmark datasets, where AVG refers to the average value of all datasets. The experimental results show that the indicator proposed in this paper has two conclusions:

\textbf{MIUB changes regularly with the change of model size.} The bolded text in Table 1 indicates that MIUB follows the pattern of decreasing as the model size increases. From the results of AVG (the average results of ARCE, ARCC, Hel, PIQA and Win), the ranks of LoRA are set to 32, 128 and 512 respectively. As the ranks increase, that is, the size of LoRA increases, MIUB gradually decreases, which means that LoRA relies less on the features of the large model and has stronger generalization. In the right half of the table, the sizes of the large model are set to $share_{8}$, $share_{4}$, $share_{3}$, $share_{1}$, which means that the parameters of every eight layers of the large model are shared, every four layers of parameters are shared, every two layers of parameters are shared, and the source large model. As the size of the large model changes, MIUB also decreases. It is worth noting that the change of the rank of the LoRA part has little effect on the model size, so the change of MIUB is small, while the change of the large model size is large, so the change of MIUB is larger.  
Additionally, as shown in Figure 2, we present the MIUB and PPL with respect to the size of the large model for the PTB and Wiki2 language modeling tasks. The experimental results indicate that, with the increase in the number of parameters, MIUB exhibits a significant decreasing trend, demonstrating that the scaling law holds.

\textbf{Compared to traditional metrics, MIUB not only better reflects the changing trend of actual effects but also exhibits greater stability in adhering to the scaling law.} 

\begin{figure}[t]
    \centering
    \includegraphics[width=0.5\textwidth]{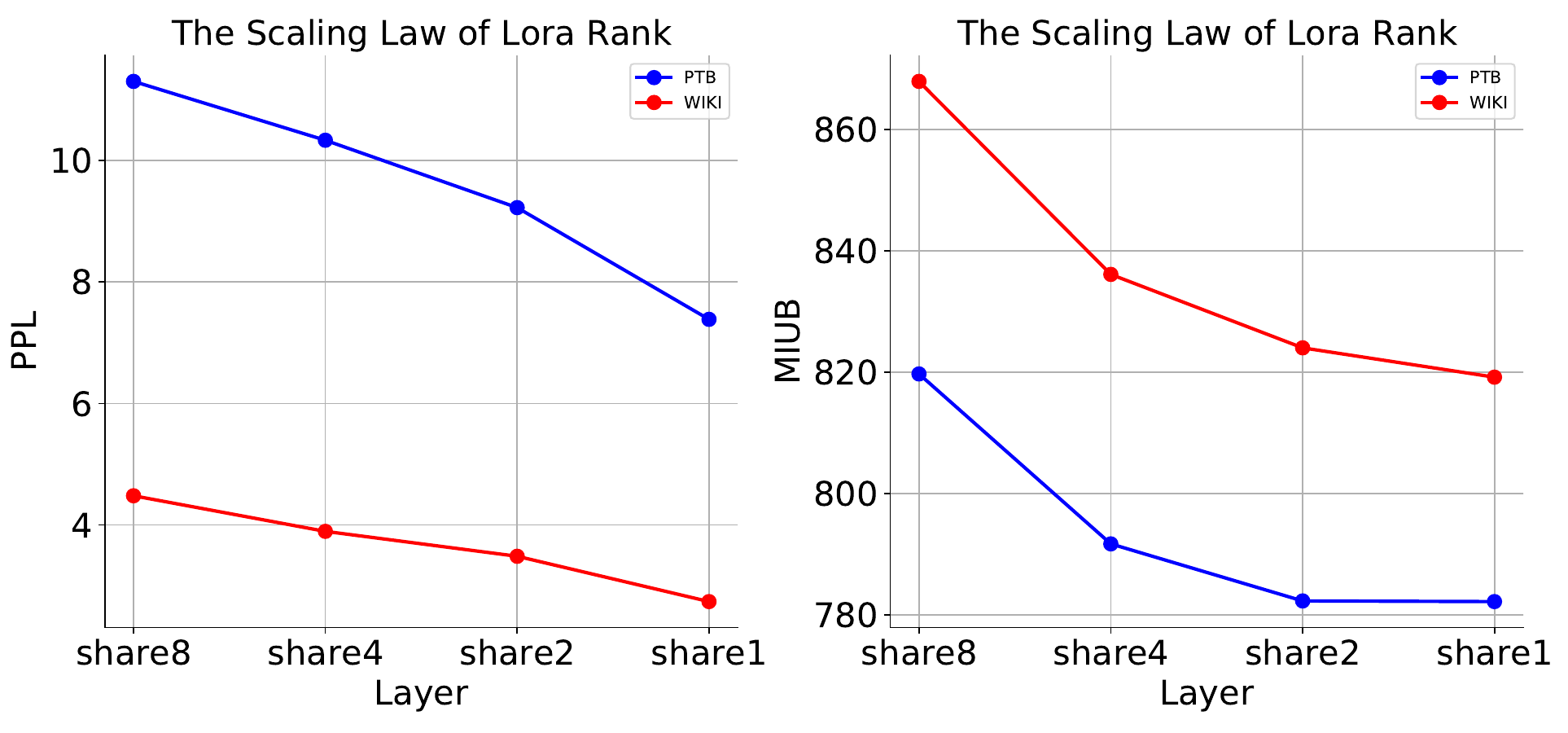} 
    \caption{Comparative experiment between MIUB and PPL under large model size changes}
    \label{fig2}
\end{figure}

We analyze this from two perspectives: first, by comparing the trends of CE, MIUB, and actual performance (ACC), and second, by conducting a comparative analysis of the scaling laws based on PPL, CE, and MIUB. As shown in Table 1, in the AVG test based on the Phi3 model, an abnormal increase in CE was observed (indicated by the red arrow) even as ACC improved. In contrast, MIUB consistently decreased as ACC increased, indicating that during fine-tuning, the model's dependency on the larger model weakened, leading to stronger learning of generalized knowledge. Regarding the stability of the scaling law, compared to CE, which exhibited a significant abnormal increase as the rank increased, MIUB consistently maintained a steady decline. As the size of the larger model increased, calculations revealed that the CE value at $share_8$ was 571 times that of $share_1$, while the size of the larger model increased by less than twice. In comparison, MIUB's change was more stable, decreasing by 17\%. This effect is more pronounced in Figure \ref{fig2}. The change in large model size is not linear; the increase becomes more significant. Additionally, under limited data conditions, although the model tends to learn more generalized knowledge, the complexity of the large model inevitably increases. Therefore, while MIUB still shows a decreasing trend, the rate of decrease will correspondingly diminish.

\subsection{Data Complexity}
\begin{figure}[t]
    \centering
    \includegraphics[width=0.5\textwidth]{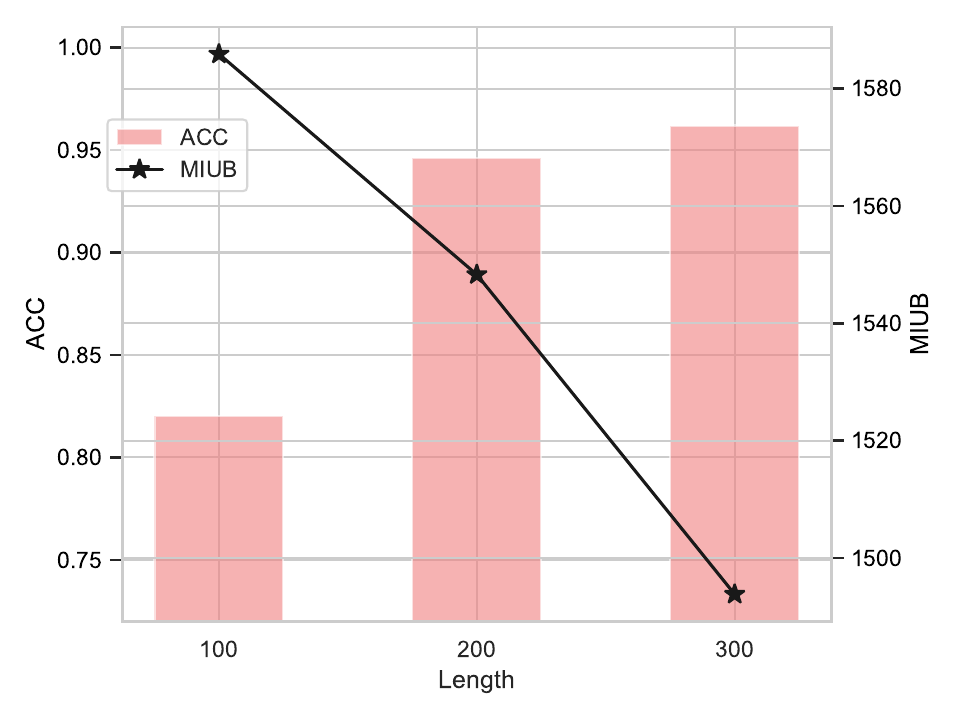} 
    \caption{The scaling law of data complexity}
    \label{fig3}
\end{figure}

To study the data size scaling law based on MIUB, we conducted experiments analyzing data complexity. As described in the Scaling Law Setting section, the data was divided according to length, and the experimental results are shown in Figure~\ref{fig3}. The results indicate that as the data length increases, the actual performance (ACC) of the large model improves, suggesting that the large model acquires more comprehensive information from the prompt. At the same time, MIUB exhibits a systematic decrease, indicating that as the new data becomes more complex (larger in scale), the dependency on the large model during fine-tuning diminishes, and there is a greater need for the LoRA module to learn more generalized knowledge.

\begin{table*}
    \centering
    \caption{The results of different prompt on Phi3-3B Model}
    \begin{tabular}{c|c|ccc|cccc}
        \toprule
         Dataset & Matrices &  32 & 128& 512 &$share_8$ &$share_4$ &$share_2$ &$share_1$\\
         \midrule
     \multirow{8}{*}{AVG}    
     & Main (ACC)  &  0.856  &0.856  & 0.864 & 0.852  &0.857 & 0.861 &0.867 \\
      &  Main (MIUB)  & \textbf{1564.6}  & \textbf{1551.5} & \textbf{1548.6} & \textbf{1894.5} & \textbf{1706.7} & \textbf{1642.2} & \textbf{1579.1}\\
                \cmidrule{2-9}
           &  Prompt1 (ACC)  & 0.854  & 0.851 & 0.854 & 0.836 & 0.832 & 0.853 & 0.865\\
            &   Prompt1 (MIUB)  & \textbf{1569.7}  & \textbf{1556.4} & \textbf{1537.4} & \textbf{1900.5} & \textbf{1704.7} & \textbf{1652.0} & \textbf{1591.0}\\
             \cmidrule{2-9}
           &   Prompt2 (ACC)  & 0.855  & 0.850 & 0.852 & 0.838 & 0.820 & 0.853 & 0.865\\
            & Prompt2 (MIUB)  & \textbf{1533.0}  & \textbf{1525.7} & \textbf{1516.1} & \textbf{1889.1}& \textbf{1682.9} & \textbf{1609.4} & \textbf{1552.0}\\
             \cmidrule{2-9}
           & Prompt3 (ACC)  & 0.869  & 0.866 & 0.870 & 0.859 & 0.856 & 0.869 & 0.872\\
            & Prompt3 (MIUB) & \textbf{1545.2}  & \textbf{1541.6} & \textbf{1534.7} & \textbf{1894.7} & \textbf{1711.0} & \textbf{1635.5} & \textbf{1573.1}\\
        \bottomrule
    \end{tabular}

    \label{tab:my_label}
\end{table*}

\subsection{MI VS MIUB}

\begin{table}
    \centering
    \caption{Comparative experiment of mutual information and upper bound of mutual information}
    \begin{tabular}{c|cc|cc}
        \toprule
        Matrices &  32 & 512 &$share_8$ &$share_1$\\
         \midrule
         ACC  & 0.870  & 0.871 &0.862  & 0.880\\
         MIUB  & 1547.121   &1533.816 &1898.560  & 1561.355\\
         MI  &  1547.120 & 1533.813 &1898.452 &1561.354 \\
        \bottomrule
    \end{tabular}

    \label{tab:my_label}
\end{table}

To compare the differences between Mutual Information (MI) and its upper bound (MIUB), this paper tests these two evaluation metrics within the LoRA fine-tuning framework. As shown in Table 2, with the increase in the rank setting of the LoRA module and the size of the large model, both MI and MIUB exhibit a downward trend, and their values are relatively close. However, as MIUB is the upper bound of MI, it consistently remains higher than MI. In certain special cases, such as small sample sizes, high model complexity, or high-dimensional data, the gap between MI and MIUB may be larger. These factors can cause MIUB to be overestimated or MI to be underestimated, thus widening the difference between the two. Therefore, MIUB is more suited for generalized scenarios.

\subsection{Prompt Learning Analysis}

In the Methodology section, this paper uses four types of prompt templates, as shown in Table 3. "Main" refers to the zero-shot prompt used for training, while "Prompt1," "Prompt2," and "Prompt3" are one-shot templates, few-shot templates with positive and negative examples, and output control templates, respectively.

The experimental results show that regardless of the template used, MIUB decreases with the increase in LoRA and the size of the large model, demonstrating the stability of MIUB as a scaling law effectiveness metric. Comparing the four prompts, the order is generally: MIUB (Prompt1) $>$ MIUB (Main) $>$ MIUB (Prompt3)  $>$  MIUB (Prompt2). Prompt1 has the highest MIUB because it incorporates data from the training set, which enhances the dependency on the large model during the knowledge learning process. In contrast, Prompt2, due to the inclusion of negative examples, has greater uncertainty and thus a smaller mutual information upper bound.

\section{Conclusion}

This paper proposes an upper bound metric for mutual information (MIUB) used for evaluating general LoRA frameworks and systematically explores the patterns of MIUB with respect to changes in large model sizes, LoRA rank sizes, and data sizes. Specifically, LoRA fine-tuning is typically a technique for enhancing the multifaceted capabilities of large models using a small amount of data and relatively few parameters in a compression context. These multifaceted capabilities may include aspects such as metaphor, multimodal, and multitasking, but in this paper, they are collectively referred to as generalization ability. Therefore, the LoRA fine-tuning process mainly involves the dependence information of large models and the newly added generalization information. Generally, stronger dependence on large models often implies weakened generalization ability, which is a key factor for the fine-tuned model's performance on new data. Leveraging the structural advantages of LoRA, this paper calculates the upper bound of mutual information (MIUB) based on the output distributions of frozen layers of LLMs and the output distributions of LoRA modules to accurately quantify the fine-tuning effects of LoRA. Experiments conducted on seven benchmark datasets and two general large models, LLaMA3-8B and Phi3-3B, show that the proposed MIUB not only aligns with the Scaling Laws of LoRA, LLM, and Data, but also provides more accurate and stable results compared to traditional general metrics such as Cross-Entropy and PPL.

\bibliography{anthology,custom}
\bibliographystyle{acl_natbib}

\end{document}